\documentclass[journal]{IEEEtran}
\usepackage{times}  
\usepackage{helvet} 
\usepackage{courier}  
\usepackage[hyphens]{url}  
\usepackage{graphicx} 
\usepackage{amsmath}
\usepackage{amssymb}
\usepackage{booktabs}
\usepackage{multirow}
\usepackage{bm}
\usepackage{verbatim}
\usepackage{xspace}
\usepackage{subfigure}
\usepackage{cite}
\usepackage{stfloats}
\usepackage{float}
\usepackage{setspace}
\usepackage[pagebackref=false,breaklinks=true,letterpaper=true,colorlinks,urlcolor=blue,citecolor=blue,linkcolor=blue,bookmarks=false]{hyperref}
\usepackage{dsfont}
\usepackage{color}
\usepackage{threeparttable}

\usepackage[capitalize]{cleveref}
\crefname{section}{Sec.}{Secs.}
\Crefname{section}{Section}{Sections}
\Crefname{table}{Table}{Tables}
\crefname{table}{Tab.}{Tabs.}

\usepackage[table]{xcolor}
\definecolor{Graylight}{gray}{0.9}

\begin{document}
%
\title{PersonMAE: Person Re-Identification Pre-Training with Masked AutoEncoders}
%
%
%

\author{Hezhen Hu,
        Xiaoyi Dong,
        Jianmin Bao,
        Dongdong Chen,
        Lu Yuan,
        Dong Chen,
        and Houqiang Li,~\IEEEmembership{Fellow,~IEEE}
\thanks{Hezhen Hu, Xiaoyi Dong, and Houqiang Li are with the University of Science and Technology of China, Hefei, 230027, China e-mail: \{alexhu, dlight\}@mail.ustc.edu.cn, lihq@ustc.edu.cn).}
\thanks{Jianmin Bao and Dong Chen are with the Microsoft Research Asia e-mail: \{jianbao, doch\}@microsoft.com.}
\thanks{Dongdong Chen and Yuan Lu are with the Microsoft Cloud AI e-mail: cddlyf@gmail.com, luyuan@microsoft.com.}
}

%
%

\markboth{IEEE Transactions on xxx,~Vol.~x, No.~x, July~2023}%
{Hu \MakeLowercase{\textit{et al.}}: PersonMAE: Person Re-Identification Pre-Training with Masked AutoEncoders}
%



\maketitle

\begin{abstract}
Pre-training is playing an increasingly important role in learning generic feature representation for Person Re-identification~(ReID).
We argue that a high-quality ReID representation should have three properties, namely, multi-level awareness, occlusion robustness, and cross-region invariance.
To this end, we propose a simple yet effective pre-training framework, namely PersonMAE, which involves two core designs into masked autoencoders to better serve the task of Person Re-ID. 
1) PersonMAE generates two regions from the given image with \textit{RegionA} as the input and \textit{RegionB} as the prediction target.
\textit{RegionA} is corrupted with block-wise masking to mimic common occlusion in ReID and its remaining visible parts are fed into the encoder.
2) Then PersonMAE aims to predict the whole \textit{RegionB} at both pixel level and semantic feature level. 
It encourages its pre-trained feature representations with the three properties mentioned above.
These properties make PersonMAE compatible with downstream Person ReID tasks, leading to state-of-the-art performance on four downstream ReID tasks, \emph{i.e.,} supervised~(holistic and occluded setting), and unsupervised~(UDA and USL setting).
Notably, on the commonly adopted supervised setting, PersonMAE with ViT-B backbone achieves $79.8\%$ and $69.5\%$ mAP on the MSMT17 and OccDuke datasets, surpassing the previous state-of-the-art by a large margin of +8.0 mAP, and +5.3 mAP, respectively.

\end{abstract}


%
\IEEEpeerreviewmaketitle

\section{Introduction}
\IEEEPARstart{P}{erson} re-identification~(ReID) aims to match a specific person across different camera views, which is a challenging retrieval problem.
Its challenges largely come from the presence of disturbing factors, \emph{e.g.} unconstrained recording conditions, occlusion, cropping misalignment, \emph{etc.}
Although current ReID methods~\cite{fu2020improving,zheng2015scalable,wang2018learning,suh2018part,wei2018person,ge2020self,yang2021learning,he2021transreid} have achieved remarkable progress, they are data-hungry and usually suffer the overfitting issue due to limited annotated ReID data.
To alleviate this issue, some methods~\cite{fu2021unsupervised,luo2021self,zhu2022pass} resort to self-supervised contrastive pre-training on large-scale unlabeled pedestrian data and demonstrate promising results.

\begin{figure}[t!]
\centering
\includegraphics[width=\linewidth]{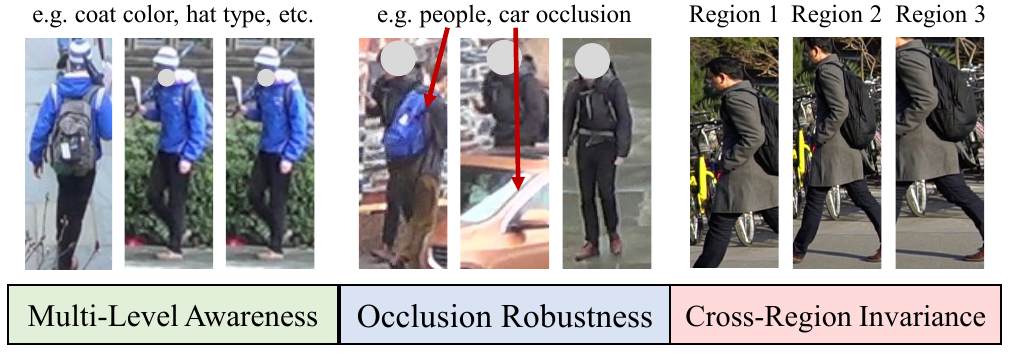}
\caption{Illustration of the main properties a good ReID representation should have, \emph{i.e.,} multi-level awareness, occlusion robustness, and cross-region invariance.
}
\label{fig:overview}
\end{figure}

However, these pre-trained feature presentations are not well-suited for person ReID. 
We argue that a good pre-training representation for ReID should have three main properties as shown in Figure~\ref{fig:overview}. 
1) \textit{\textbf{Multi-Level Awareness.}} Pedestrians with quite similar wearing exist commonly in ReID tasks, so the visual representation should be aware of all the discriminative cues adequately, including both the semantics (a broad range of aspects, such as person clothes, belongings and their context) and low-level details (such as the color of the bag).
2) \textit{\textbf{Occlusion Robustness.}} In real-world scenarios, the target pedestrian is often occluded by surrounding objects, such as cars, trees, non-target people, \emph{etc.} So the visual representation should be robust to diverse occlusion and focus on the key cues of the target pedestrian.
3) \textit{\textbf{Cross-Region Invariance.}} 
Pedestrian images are usually cropped via the off-the-shelf human detector/tracker, which leads to misalignment issues.
The visual representation of the same pedestrian should be invariant, regardless of cropping misalignment.

Previous self-supervised ReID pre-training ~\cite{fu2021unsupervised,luo2021self,zhu2022pass} methods are dominated by contrastive learning~(CL).
They generally learn semantics by pulling the global features of two augmented regions of the same image closer and pushing the features of two different images away. 
While their pre-trained features inherently satisfy the cross-region invariance property, they are deficient in the remaining two properties. 
Specifically, for occlusion robustness, although random erasing is used as one augmentation in contrastive pre-training, their used mask ratios are often relatively small, resulting in their representations that cannot handle large occlusion well.
Additionally, since the contrastive loss is only applied upon the global feature of the image, and lacks supervision on each token, it yields a bottleneck in capturing low-level and local detail information.

In this paper, we aim to satisfy three important properties and build a new pre-training framework, namely PersonMAE.
It introduces cross-region multi-level prediction objective, in coordination with masked autoencoders~\cite{he2022masked}.
Specifically, with a given pedestrian image, we first generate two regions with the cross-region generation module, \emph{i.e.,} the input \textit{RegionA} and prediction target \textit{RegionB}.
Then we mask a large portion of \textit{RegionA} and only feed its visible parts into the encoder.
Finally, the decoder leverages the encoder output and the cross-region relationship between \textit{RegionA} and \textit{RegionB} to predict the whole \textit{RegionB} at both low-level pixel space and high-level semantic space.

Our PersonMAE framework differs substantially from previous contrastive-based approaches in its emphasis on the reasoning capability of the whole human image to better serve the ReID task.
It is achieved by: 1) incorporating per-token supervision on both low-level pixel and high-level semantics, thereby explicitly guiding the model's learning of \textit{\textbf{multi-level awareness}}.
2) The input region image is heavily masked and only a small visible part is passed to the encoder, compelling it to capture all relevant features, not merely the most prominent ones in the full image.
This ``mask and drop'' operation facilitates the model's learning of \textit{\textbf{occlusion robustness}}.
3) The pseudo-ground-truth is provided by another region, which simulates the effects of jittering human detection.
The model embeds \textit{\textbf{cross-region invariance}} through an outer prediction approach that exploits the consistency between these two regions.

We pre-train our PersonMAE on the widely used LUPerson~\cite{fu2021unsupervised} dataset and thoroughly evaluate the performance of PersonMAE under four settings, \emph{i.e.,} supervised holistic and occluded ReID, unsupervised domain adaptation~(UDA), and unsupervised learning~(USL) setting.
For the most commonly used supervised settings, we reach new SOTA results with $79.8$ mAP on the challenging MSMT17 and $69.5$ mAP on the OccDuke dataset, surpassing the previous SOTA method PASS~\cite{zhu2022pass} with $+8.0$ and $+5.3$ respectively. 
When it comes to the unsupervised settings, we get $53.0$ and $48.8$ mAP on MSMT17 for UDA and USL ReID, outperforming PASS by $+4.1$ and $+7.8$ respectively.

Our contributions are summarized as follows,
\begin{itemize}
    \item We analyze the challenges of the ReID tasks and identify three important properties that a desired ReID representation should satisfy: multi-level awareness, occlusion robustness, and cross-region invariance.
    \item To achieve this goal, we propose a simple yet effective ReID pre-training framework called PersonMAE. PersonMAE introduces a cross-region multi-level prediction objective, in cooperation with masked autoencoders to exhibit these desired properties.
    \item Through extensive experiments, our PersonMAE demonstrates superior performance, achieving SOTA performance in a series of downstream ReID tasks on different datasets.
\end{itemize}

\section{Related Work}
\subsection{Person Re-Identification}
Person re-identification aims to match a specific person across different camera views.
It is a kind of retrieval problem, whose challenges come from unconstrained recording conditions, occlusion, cropping errors, \emph{etc.}
Generally, ReID contains supervised and unsupervised task settings, which are introduced as follows.

\noindent \textbf{Supervised Person ReID.}
Typically, it contains two settings, \emph{i.e.,} holistic and occluded ReID.
Holistic ReID is a general fully-supervised setting and aims to retrieve a certain person across different camera views.
Current methods have achieved remarkable progress~\cite{ye2021deep,li2014deepreid,wang2016joint,chen2019self,kalayeh2018human,zheng2019joint,yang2022domain,chen2020salience,li2021combined,park2020relation,yu2019robust,zhang2020relation,yan2022beyong}, and can be grouped into two mainstreams.
One focuses on the design of novel architectures to mine fine-grained discriminative cues~\cite{sun2018beyond,suh2018part,wang2018learning,guo2019beyond,zhou2019omni,ren2022s2net,hou2019interaction,he2021transreid}.
MGN~\cite{wang2018learning} emphasizes the fine-grained details by explicitly splitting the holistic pedestrian image into multiple granularities.
TransReID~\cite{he2021transreid} organizes the holistic image as multiple local tokens and first leverages the strong Vision Transformer~(ViT)~\cite{dosovitskiy2021an}.
Another mainstream resorts to effective optimization, including the hard triplet loss~\cite{hermans2017defense}, circle loss~\cite{sun2020circle}, rank fusion~\cite{paisitkriangkrai2015learning}, \emph{etc.}

Different from holistic ReID, the occluded counterpart is a harder task and its query set is constructed by occluded pedestrian images~\cite{peng2022deep,zhuo2018occluded, he2019foreground, huang2020human, wang2022feature}.
Zhuo~\emph{et al.}~\cite{zhuo2018occluded} first study this problem via the attention mechanism and occlusion simulation.
The following works further disentangle discriminative cues from the occlusion with the assistance of pose landmarks~\cite{he2019foreground,he2020guided} or human parsing~\cite{huang2020human,yu2021neighbourhood}.

\noindent \textbf{Unsupervised Person ReID.}
This task directly trains a model without utilizing annotation of the target dataset~\cite{lin2019bottom,fan2018unsupervised,fu2019self,ge2020mutual,ge2020self,dai2021cluster,zhang2022implicit,yang2020leveraging,cho2022part}.
There are two typical categories, \emph{i.e.,} unsupervised domain adaptation~(UDA) and unsupervised learning~(USL).
SpCL~\cite{ge2020self} proposes a self-paced contrastive learning framework, which gradually fine-tunes the network with the pseudo labels generated by reliable clustering.
As one of the state-of-the-art methods, C-Contrast~\cite{dai2021cluster} computes the contrastive loss based on the cluster-level contrastive loss to relieve the inconsistent updating progress among different clusters.

\begin{figure*}[t!]
\centering
\includegraphics[width=0.9\linewidth]{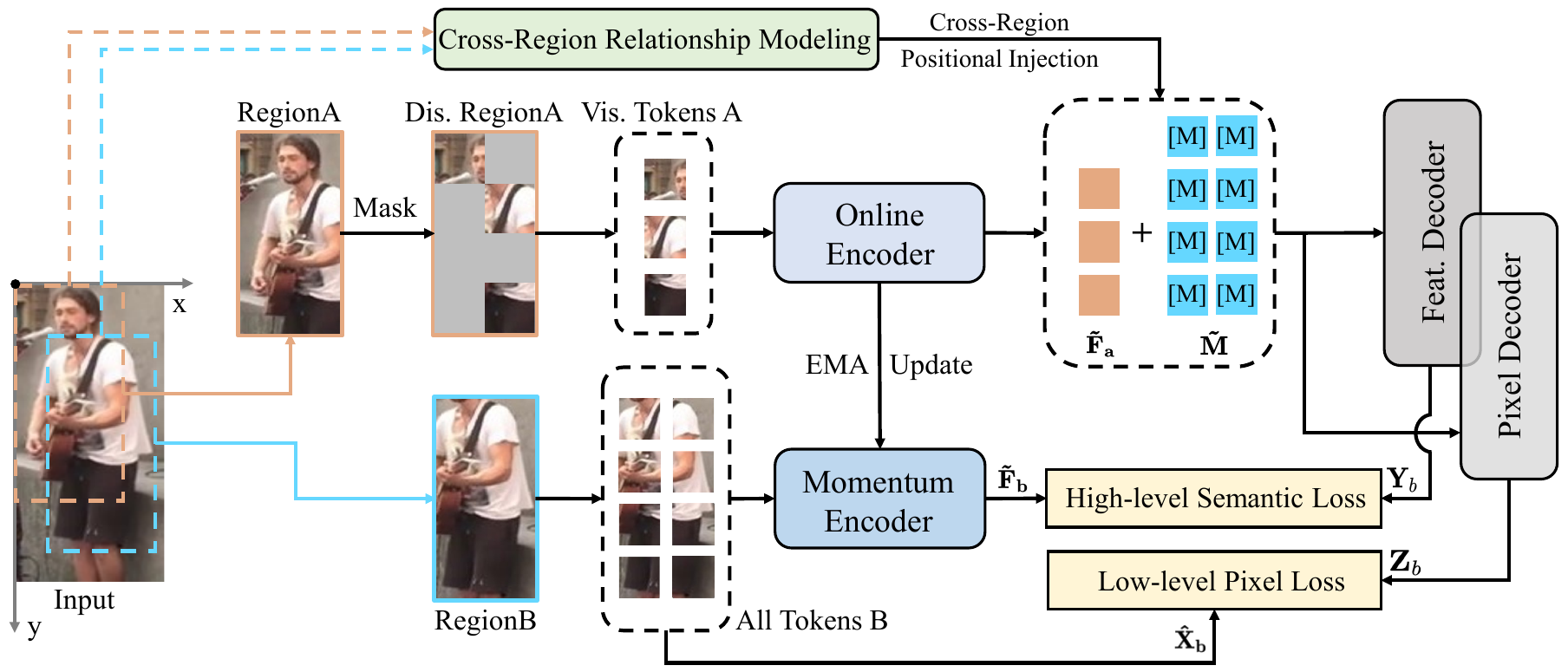}
\caption{Illustration of our PersonMAE pre-training framework. 
Given the global person image, PersonMAE utilizes the cross-region generation module for two regions, \emph{i.e.,} the input \textit{RegionA} and prediction target \textit{RegionB}.
\textit{RegionA} is distorted with masking, and its remaining part is fed into the encoder.
The decoder leverages the encoder output and their cross-region relationship to perform the whole prediction on both low-level clues and high-level semantics of \textit{RegionB}.
}
\label{fig:framework}
\end{figure*}

\subsection{Self-Supervised Pre-Training}
Self-supervised pre-training aims to learn more generic feature representations from a large amount of unlabeled data, which benefits the downstream tasks.
Among them, contrastive learning has been widely studied for pre-training~\cite{he2020momentum,chen2020simple,caron2021emerging,li2021improve,li2021unsupervised}, which aims to pull the representation of similar instances closer, while pushing away negative instances.
In order to obtain informative negative instances, some works resort to the techniques like memory banks~\cite{he2020momentum} and large batch sizes~\cite{chen2020simple}.
There also exist some other works~\cite{grill2020bootstrap,chen2021exploring} achieving great performance without the requirement of negative instance.
DINO~\cite{caron2021emerging} achieves overwhelming performance by emphasizing the distillation between global and local views.
Recently, motivated by the success of BERT~\cite{devlin2019bert} in NLP, some works~\cite{he2022masked,bao2022beit,dong2022bootstrapped,dong2021peco,wang2022bevt,huang2022contrastive,tao2022siamese,xie2022simmim} works with masked autoencoders~(MAE) and achieve promising performance.
Typically, MAE adopts the ViT backbone and aims to predict the masked patches from the remaining visible ones.

In person ReID, there exist works leveraging the success of self-supervised pre-training~\cite{fu2021unsupervised,yang2022unleashing,fu2022large,zhu2022pass}.
Fu~\emph{et al.}~\cite{fu2021unsupervised} proposes a large-scale LUPerson dataset and pioneers the exploration of contrastive-based pre-training based on this dataset.
Luo~\emph{et al.}~\cite{luo2021self} investigate contrastive-based pre-training based on ViT and improve pre-training via reducing the domain gap between pre-training and fine-tuning data.
PASS~\cite{zhu2022pass} further proposes part-aware pre-training based on DINO.
These methods demonstrate their effectiveness on the downstream ReID tasks and are all based on contrastive learning.
Different from them, we aim to make the first attempt to leverage the advance of masked autoencoders into ReID pre-training.

\section{Methodology}

\subsection{Overview}
As we briefly introduced in Section 1, a good person-ReID model should maintain three important properties: multi-level awareness, occlusion robustness, and cross-region invariance. 
To maintain these properties, we propose our PersonMAE framework.
As shown in Figure~\ref{fig:framework}, with a given image, we first generate the input image \textit{RegionA} and prediction target \textit{RegionB} with a cross-region generation module.
Then we mask a large portion of \textit{RegionA} and use the unmasked part as the online encoder input. 
The decoders use both the encoder output features and mask tokens to predict the whole \textit{RegionB}. 
As is well known, both low-level clues (such as the color of clothes and hair) and high-level semantics (such as global identity information) matter in ReID.
Therefore, we utilize a pixel decoder to predict the raw pixels of \textit{RegionB} for low-level reasoning, and a feature decoder to predict the feature of \textit{RegionB} for high-level semantic learning.

\subsection{Cross-Region Generation}
Previous general MAE-based pre-training methods always mask part of the input and then predict itself.
It is an \textit{inner prediction} manner that aims to model and understand the semantic content within the input image. 
However, this prediction manner may hardly handle the disturbance from unreliable cropping of pedestrian images.
To help the model learn robust representation regardless of such misalignment, we propose our cross-region generation module.

In detail, with a given pedestrian image, we first perform random resizing as follows,
\begin{equation}
  \mathbf{\tilde{X}} = Ip(\mathbf{X}, [H+p, W+p/2]), \ s.t. \ p \in [0, m],
\end{equation}
where $Ip(\cdot, \cdot)$ denotes the interpolation function, whose inputs are the original image and desired output image size.
$m$ represents the maximum shift size.
With the resized image $\mathbf{\tilde{X}}$, we extract two regions at the same scale with the size of $[H, W]$ as follows,
\begin{equation}
\begin{gathered}
  \mathbf{\tilde{X}_a} = \mathbf{\tilde{X}}[(s_h^a, s_w^a), (H, W)], \\
  \mathbf{\tilde{X}_b} = \mathbf{\tilde{X}}[(s_h^b, s_w^b), (H, W)],
\end{gathered}
\end{equation}
where $s_h^a = s_w^a = 0$, $s_h^b \in [0, p]$ and $\ s_w^b \in [0, p/2]$ and these two tuples represent the coordinate of the upper-left corner and cropped image size, respectively.
Then these two regions are separately split into non-overlapping patches, whose total number is $N= H \times W / T^2 $ and $T$ represents the patch size.
With this step, a certain region is transformed into the visual token sequence, \emph{e.g.} $ \mathbf{\tilde{X}_a} = {\{\bm{x}^i_a\}}^N_{i=1}, \bm{x}_a^i \in \mathbf{R}^{CT^2}$. $\bm{x}_a^i$ is obtained via squeezing the visual token and $C$ represents the image channel dimension.
The same can be obtained for $\mathbf{\tilde{X}_b}$.

It is an \textit{outer prediction} problem that the model should not only understand the content of the input \textit{RegionA}, but also the invariance between \textit{RegionA} and \textit{RegionB}.
It is much more challenging than previous \textit{inner prediction} and better improves the cropping-error robustness of the model.

\subsection{Semantic Extraction in Encoder}
Occlusion is one of the most challenging problems in the ReID task, since the pedestrian may be occluded by some obstacles, \emph{e.g.} trees, cars, walls, and other passengers.
So it requires the model not only focus on the most salient feature of the input but all the reasonable features within it. 
To mimic various occlusions during training, we utilize the block-wise masking strategy introduced in BEiT~\cite{bao2022beit} for the input \textit{RegionA} and only use the unmasked part as the encoder input. 
Formally:
\begin{equation}
\begin{gathered}
  \mathcal{M} = \{m_1, ..., m_N\}, \; m_i \in \{0,1\},
  \ s.t.  \sum{\mathcal{M}} = r \times N,\\
  \mathbf{\tilde{F}_a} = E_o(\mathbf{{V}_a}), \; \mathbf{{V}_a} = \{x_a^i|m_i=0 \}
\end{gathered}
\end{equation}
where $\mathcal{M}$ is the sampling function without replacement and the number of the masked positions is $r \times N$. $E_o(\cdot)$ denotes the online encoders with parameters $\theta_o$.

\subsection{Cross-Region Relationship Modeling}
For previous inner prediction methods, the input and target images share the region, so they could use the same position information directly. On the contrary, our outer prediction design leads to a different region for input and target, the decoder needs to predict the target \textit{RegionB} based on the feature of \textit{RegionA} and their cross-region relationship. 

In practice, we use absolute position embedding to model such a relationship. We build a coordinate system and set the upper-left corner of the \textit{RegionA} as the origin.
Denote $x$ and $y$ are the horizontal and vertical index of the predicted region. Formally:
\begin{equation}
\begin{gathered}
  r(x, y) = (x +  \frac{s_h^b - s_h^a}{T}, \ y + \frac{s_w^b - s_w^a}{T}),
\end{gathered}
\end{equation}
where $x \in [0, H/T-1]$ and $y \in [0, W/T-1]$.
We follow common practices to convert $r(x,y)$ into the fixed 2D position embedding $\mathbf{P}_r$ based on $sin(\cdot)$ and $cos(\cdot)$ operators~\cite{he2022masked}.

Then the formulated shifting relation and the output $\mathbf{\tilde{F}_a}$ of the online encoder are jointly fed into the following two decoders for \textit{RegionB} prediction, \emph{i.e.,} pixel regression decoder and feature prediction decoder.

\subsection{Low-Level Reasoning via Pixel Regression}
As illustrated above, low-level clues are important for the ReID task. Therefore, a pixel decoder is adopted to conduct prediction at the low-level aspect via recovering all \textit{RegionB} pixels.
Specifically, we organize $N$ learnable tokens as the mask tokens $\mathbf{M}_p$.
To inform the framework where the corresponding output of each mask token should predict, we further add the position embedding $\mathbf{P}_r$ on $\mathbf{M}_p$, denoted as $\mathbf{\tilde{M}}_p$.
Then the PE-informed $\mathbf{\tilde{M}}_p$ is concatenated with the normalized $\mathbf{\tilde{F}_a}$ as the input of the pixel decoder.
The pixel decoder contains two ViT blocks, followed by an MLP layer to regress the missing pixel value.
Denote the output of the pixel decoder as $\mathbf{Z}_b=\{\bm{z}^i_b|i \in [0, N-1]\}$.
During pre-training, the low-level region-consistency constraint is calculated as follows,
\begin{equation}
\begin{gathered}
  \hat{\bm{x}}^i_b = (\bm{x}^i_b - m^i_b) / s^i_b, \ i = 0, ..., N-1 \\
  \mathcal{L}_p = \sum_{i=1}^N \frac{1}{T^2C} || \hat{\bm{x}}^i_b - \bm{z}^i_b ||_2^2,
\end{gathered}
\end{equation}
where $m^i_b$ and $s^i_b$ are the mean and standard deviation of the $i$-th patch and we utilize the normalized patch as the target.
This constraint serves as a strong regularization term during optimization and makes the framework model human statistics via reasoning about the low-level textures.

\subsection{High-Level Understanding via Feature Prediction}
Meanwhile, high-level semantics are the core for person re-identification, so we use the feature decoder to conduct feature prediction.
Similar to the pixel regression branch, we adopt a new set of $N$ learnable tokens as the mask token $\mathbf{M}_f$ and associate it with the same position embedding $\mathbf{P}_r$, denoted as $\mathbf{\tilde{M}}_f$.
It is concatenated with the normalized $\mathbf{\tilde{F}_a}$ as the input of the feature decoder.
The feature decoder contains two ViT blocks, followed by a fully-connected layer to project the features on the desired semantic space. 
We argue that dense local supervision forces the model to capture more fine-grained details, so the prediction target is all the local features of \textit{RegionB} encoded by an EMA model. 
Formally,
\begin{equation}
\label{equ:ema}
\begin{gathered}
  \mathbf{\tilde{F}_b} = E_m(\mathbf{{V}_b}), \mathbf{{V}_b} = \mathbf{\tilde{X}_b}  \\
  \bm{\theta}_m \leftarrow \gamma\bm{\theta}_m + (1-\gamma) \bm{\theta_o},
\end{gathered}
\end{equation}
where $E_m(\cdot)$ is the momentum encoder, $\theta_o$ and $\theta_m$ represent the parameters of the online encoder and momentum encoder, respectively. 
Compared with the online encoder $E_o$, the parameters of $E_m$ are updated much slower and provide a more stable prediction target, which is beneficial for modeling the pedestrian statistics in the semantic space.

With the prediction target $\mathbf{\tilde{F}_b}=\{\bm{f}^i_b|i \in [0, N-1]\}$, we denote the output of feature prediction decoder as $\mathbf{Y}_b=\{\bm{y}^i_b|i \in [0, N-1]\}$ and the feature-level region consistency constraint is formulated as follows,
\begin{equation}
\label{equ:smooth}
\begin{gathered}
  \mathcal{L}_f = \sum_{i=1}^N \texttt{SmoothL1}\left(\bm{y}^i_b,\  \texttt{SG}(\bm{f}^i_b),\  \beta \right),
\end{gathered}
\end{equation}
where we utilize smooth L1 loss for feature matching.
$\texttt{SG}(\cdot)$ represents the stop gradient function and $\beta$ is the smoothing factor.
Overall, during pre-training, the objective function is formulated as follows,
\begin{equation}
\begin{gathered}
  \mathcal{L} = \mathcal{L}_p + \lambda\mathcal{L}_f,
\end{gathered}
\end{equation}
where $\lambda$ is the weighting factor.
$\mathcal{L}_p$ and $\mathcal{L}_f$ are the low-level and semantic constraint, respectively.

\section{Experiment}

\subsection{Evaluation Protocol}

\noindent \textbf{Datasets.}
We pre-train our PersonMAE on the large-scale pedestrian dataset LUPerson~\cite{fu2021unsupervised}.
It contains 4.18M unlabeled images with 46,260 diverse scenes, which are collected from the Internet.
For downstream ReID tasks, we conduct experiments on three datasets, \emph{i.e.,} Market1501~\cite{zheng2015scalable}, MSMT17~\cite{wei2018person} and OccDuke~\cite{miao2019pose}.
Market1501 and MSMT17 are general holistic person ReID benchmarks, which contain 32,668 images from 1,501 persons and 126,441 images from 4,101 persons, respectively.
OccDuke contains 35,489 images from 1,812 persons, which is the most challenging occluded person ReID benchmark.

\noindent \textbf{Evaluation metrics.}
Following common practices, we utilize the cumulative matching characteristics at Top-1~(Rank1) and mean average precision~(mAP) for performance evaluation.

\noindent \textbf{Implementation details.}
We introduce the specific settings during pre-training, supervised ReID and unsupervised ReID settings.

\emph{Pre-training setting.}
Pre-training is conducted on 8$\times$V100 GPUs for 100 epochs. 
Adam is adopted for optimization.
The learning rate is set to 1.2e-3, with a warmup of 20 epochs, and single-cycle cosine learning rate decay.
The weight decay is set as 0.05. 
We only utilize horizontal flipping and random cropping on the raw data to generate the global pedestrian image.
The input resolution is 256$\times$128.
$\gamma$ represents momentum coefficient.
It is initially set as 0.999, linearly changed to 0.9999 in the first 20 epochs, and kept as 0.9999 for the remaining epochs.
$\beta$ represents the smoothing factor and is set to 2.

\emph{Supervised ReID setting.}
During fine-tuning, we feed the unmasked image into the pre-trained online encoder for further optimization.
We follow the common practice in~\cite{fu2021unsupervised} and utilize the MGN~\cite{wang2018learning} head for downstream ReID tasks.
During supervised fine-tuning, we utilize the layer-wise lr decay strategy following~\cite{he2022masked,bao2022beit}.
The batch size is set as 256, with 16 pedestrians in each batch.
Adam optimizer is adopted with the learning rate and weight decay set as 5e-3 and 0.05, respectively.
The layer decay is set as 0.65.
The training lasts 60 epochs, and we utilize the cosine learning rate decay scheduler.
We keep the same setting for supervised holistic and occluded ReID.

\begin{table}[t!]
\small
\tabcolsep=1.6pt
\begin{center}
\caption{Comparison with state-of-the-art methods on holistic supervised ReID.
We conduct experiments on Market1501 and MSMT17 datasets.
* means the result with the input size as $384\times128$.
$^{\dag}$ represents our implementation.
}
\begin{threeparttable}
\begin{tabular}{llcccc}
\toprule[1pt]
\multirow{2}{*}{\textbf{Method}} & \multirow{2}{*}{\textbf{Backbone}} & \multicolumn{2}{c}{\textbf{Market1501}}  & \multicolumn{2}{c}{\textbf{MSMT17}}  \\ \cline{3-6}
       &  & \textbf{mAP} & \textbf{Rank1} & \textbf{mAP} & \textbf{Rank1} \\ 
       \hline 
\multicolumn{6}{l}{\textit{Pretraining on ImageNet1K-1.3M}} \\
PCB~\cite{sun2018beyond}~(2018) & Res-50 & 81.6 & 93.8  & - & - \\
MGN*~\cite{wang2018learning}~(2018) & Res-50 & 87.5 & 95.1 & 63.7 & 85.1  \\
BOT~\cite{luo2019strong}~(2019) & Res-50 & 85.9 & 94.5 & 50.2 & 74.1 \\
ABDNet*~\cite{chen2019abd}~(2019) & Res-50 & 88.3 & 95.6 & 60.8 & 82.3 \\
SAN~\cite{jin2020semantics}~(2020) & Res-50 & 88.0 & 96.1 & 55.7 & 79.2 \\
GASM~\cite{he2020guided}~(2020) & Res-50 & 84.7 & 95.3 & 52.5 & 79.5 \\
ISP~\cite{zhu2020identity}~(2020) & Res-50 & 88.6 & 95.3 & - & - \\
HOReID~\cite{wang2020high}~(2020) & Res-50 & 84.9 & 94.2 & - & - \\
FA-Net~\cite{liu2021end}~(2021) & Res-50 & 84.6 & 95.0 & 51.0 & 76.8 \\
PAT~\cite{li2021diverse}~(2021) & Res-50 & 88.0 & 95.4 & - & - \\
NFormer~\cite{wang2022nformer}~(2022) & Res-50 & 91.1 & 94.7 & 59.8 & 77.3 \\
\hline
\multicolumn{6}{l}{\textit{Pretraining on ImageNet21K-14M}} \\
TransReID~\cite{he2021transreid}~(2021) & ViT-B & 87.4 & 94.6 & 63.6 & 82.5 \\
AAformer*~\cite{zhu2021aaformer}~(2021) & ViT-B & 87.7 & 95.4 & 63.2 & 83.6 \\
FED~\cite{wang2022feature}~(2022) & ViT-B & 86.3 & 95.0 & - & - \\
DCAL~\cite{zhu2022dual}~(2022) & ViT-B & 87.5 & 94.7 & 64.0 & 83.1 \\
\hline
\multicolumn{6}{l}{\textit{Pretraining on LUPerson-4.2M}} \\
MoCov2*~\cite{chen2020improved,fu2021unsupervised}~(2021) & Res-50 & 91.0 & 96.4 & 65.7 & 85.5 \\
UPReID~\cite{yang2022unleashing}~(2022) & Res-50 & 91.1 & 97.1 & 63.3 & 84.3 \\ 
MoCov3~\cite{chen2021empirical,luo2021self}~(2021) & ViT-S & 82.2 & 92.1 & 47.4 & 70.3 \\
MoBY~\cite{xie2021self,luo2021self}~(2021) & ViT-S & 84.0 & 92.9 & 50.0 & 73.2 \\
DINO~\cite{caron2021emerging,luo2021self}~(2021) & ViT-S & 90.3 & 95.4 & 64.2 & 83.4 \\
DINO-CFS~\cite{luo2021self}~(2021) & ViT-S & 91.0 & 96.0 & 66.1 & 84.6 \\
MAE$^{\dag}$~\cite{he2022masked}~(2022) & ViT-B & 91.1 & 96.1 & 71.2 & 88.0 \\
PASS~\cite{zhu2022pass}~(2022) & ViT-S & 92.2 & 96.3 & 69.1 & 86.5 \\
PASS~\cite{zhu2022pass}~(2022) & ViT-B & 93.0 & 96.8 & 71.8 & 88.2 \\
\rowcolor{Graylight} 
Ours & ViT-S & \textbf{92.5} & \textbf{96.7} & \textbf{75.2} & \textbf{89.1} \\
\rowcolor{Graylight} 
Ours & ViT-B & \textbf{93.6} & \textbf{97.1} & \textbf{79.8} & \textbf{91.4} \\
\bottomrule[1pt]
\end{tabular}
\end{threeparttable}
\end{center}
\label{tab:supervised_reid}
\end{table}

\emph{Unsupervised ReID setting.}
For unsupervised ReID tasks, we build upon C-Contrast~\cite{dai2021cluster, luo2021self} with the ViT backbone and follow most of its settings.
UDA and USL ReID only differ in model initialization.
UDA utilizes the pre-trained model from the source dataset, while USL directly utilizes our LUPerson pre-trained parameters.
Then the training is conducted in an unsupervised manner on the target dataset.
The visual features corresponding to images in the training set are first extracted, followed by clustering for pseudo labels and corresponding clustering centers.
Then the framework leverages the contrastive objective between output features and clustering centers during optimization.
More details can be found in~\cite{dai2021cluster}.
SGD optimizer is adopted.
The initial learning rate is set as 0.007 and is decayed with 0.1 every 20 epochs.
The batch size is set as 256, with 32 pedestrians in each batch.

\subsection{Comparison with State-of-the-Art Methods}
We compare our method with previous state-of-the-art methods on four different downstream ReID tasks, \emph{i.e.,} supervised~(holistic and occluded ReID), and unsupervised~(UDA and USL ReID).
Note that we do not adopt any post-processing method like Re-Rank~\cite{zhong2017re} in our method.

\begin{table}[t!]
\small
\tabcolsep=7.6pt
\begin{center}
\caption{Comparison with state-of-the-art methods on occluded supervised ReID. Experiments are conducted on OccDuke. $^{\dag}$ denotes our implementation.}
\begin{threeparttable}
\begin{tabular}{llcc}
\toprule[1pt]
\multirow{2}{*}{\textbf{Method}} & \multirow{2}{*}{\textbf{Backbone}} & \multicolumn{2}{c}{\textbf{OccDuke}} \\ \cline{3-4}
       &  & \textbf{mAP} & \textbf{Rank1} \\ 
       \hline
\multicolumn{4}{l}{\textit{Pretraining on ImageNet1K-1.3M}} \\
PCB~\cite{sun2018beyond}~(2018) & Res-50 & 33.7 & 42.6 \\
DSR~\cite{he2018deep}~(2018) & Res-50 & 30.4 & 40.8 \\
Ad-Occ~\cite{huang2018adversarially}~(2018) & Res-50 & 32.2 & 44.5 \\
PGFA~\cite{miao2019pose}~(2019) & Res-50 & 37.3 & 51.4 \\
RE~\cite{zhong2020random}~(2020) & Res-50 & 30.0 & 40.5 \\
PVPM~\cite{gao2020pose}~(2020) & Res-50 & 37.7 & 47.0 \\
ISP~\cite{zhu2020identity}~(2020) & Res-50 & 52.3 & 62.8 \\
HOReID~\cite{wang2020high}~(2020) & Res-50 & 43.8 & 55.1 \\
PAT~\cite{li2021diverse}~(2021) & Res-50 & 53.6 & 64.5 \\
LKWS~\cite{yang2021learning}~(2021) & Res-50 & 46.3 & 62.2 \\
\hline
\multicolumn{4}{l}{\textit{Pretraining on ImageNet21K-14M}} \\
TransReID~\cite{he2021transreid}~(2021) & ViT-B & 53.1 & 60.5 \\
PFD~\cite{wang2022pose}~(2022) & ViT-B & 60.1 & 67.7 \\
FED~\cite{wang2022feature}~(2022) & ViT-B & 56.4 & 68.1 \\ \hline
\multicolumn{4}{l}{\textit{Pretraining on LUPerson-4.2M}} \\
DINO-CFS*~\cite{luo2021self}~(2021) & ViT-S & 58.3 & 67.7 \\
MAE$^{\dag}$~\cite{he2022masked}~(2022) & ViT-B & 60.8 & 68.1 \\
PASS$^{\dag}$~\cite{zhu2022pass}~(2022) & ViT-S & 59.7 & 69.0  \\
PASS$^{\dag}$~\cite{zhu2022pass}~(2022) & ViT-B & 64.2 & 74.5 \\
\rowcolor{Graylight} 
Ours & ViT-S & \textbf{65.2} & \textbf{72.0} \\
\rowcolor{Graylight} 
Ours & ViT-B & \textbf{69.5} & \textbf{76.0} \\
\bottomrule[1pt]
\end{tabular}
\end{threeparttable}
\end{center}
\label{tab:supervised_occreid}
\end{table}

\noindent \textbf{Supervised ReID.}
As shown in Table~\ref{tab:supervised_reid}, we generally divide previous methods by model initialization, \emph{i.e.,} utilizing supervised ImageNet pre-training and self-supervised LUPerson pre-training as backbone initialization, respectively.
For methods adopting ImageNet-supervised pre-trained parameters, they usually focus on designing novel architectures to capture fine-grained pedestrian details.
With the popularity of Transformer, more and more current methods have turned to ViT backbone and achieved promising results.
However, they need to pre-train on ImageNet-21k dataset~(larger than LUPerson, 14M vs 4.2M data volume)~\cite{zhu2021aaformer,he2021transreid,wang2022feature,zhu2022dual}.
Compared with these works, our method pre-trains on the LUPerson dataset and achieves remarkable performance gains over the most challenging competitor, \emph{i.e.,} 5.9\% and 15.8\% mAP gain on Market1501 and MSMT17, respectively.

Methods in the bottom part conduct self-supervised pre-training on LUPerson to initialize the backbone.
Regardless of the adopted backbones, their pre-training methods are dominated by the variants of contrastive learning.
UPReID~\cite{yang2022unleashing} and PASS~\cite{zhu2022pass} design subtle techniques to further mine the information between global image and local patches.
Our method achieves new state-of-the-art performance, \emph{i.e.,} 93.6 and 79.8 mAP on Martket1501 and MSMT17 datasets, respectively.
It is worth mentioning that a more significant performance gain is achieved on more challenging MSMT17, \emph{i.e.,} 6.1\% and 8\% mAP gain on ViT-S and ViT-B over previous SOTA, respectively.

Supervised occluded ReID is evaluated on OccDuke benchmark as shown in Table~\ref{tab:supervised_occreid}.
It is a harder setting than the holistic counterpart since the given query pedestrian contains diverse occlusion conditions.
Previous methods for this setting mainly leverage external cues or design novel architectures for feature disentanglement.
ISP~\cite{zhu2020identity} iteratively learns feature maps at the pixel level to provide identity-guided parsing cues.
FED~\cite{wang2022feature} builds on the ViT-B backbone and designs two modules to eliminate the disturbance from occlusion.
Compared with previous works, our method demonstrates superiority without utilizing external cues, \emph{i.e.,} achieving 69.5\% mAP on OccDuke.
It can be attributed to our explicit occlusion-aware design in the pre-training strategy, which makes our framework model pedestrian identity statistics well even under severe occlusion.

\begin{table}[t!]
\small
\tabcolsep=2pt
\begin{center}
\caption{Comparison with state-of-the-art methods on UDA ReID. ``MS'' and ``Mar'' denote the MSMT17 and Market1501 dataset, respectively.}
\begin{threeparttable}
\begin{tabular}{llcccc}
\toprule[1pt]
\multirow{2}{*}{\textbf{Method}} & \multirow{2}{*}{\textbf{Backbone}} & \multicolumn{2}{c}{\textbf{MS$\rightarrow$Mar}}  & \multicolumn{2}{c}{\textbf{Mar$\rightarrow$MS}}  \\ \cline{3-6}
       &  & \textbf{mAP} & \textbf{Rank1} & \textbf{mAP} & \textbf{Rank1} \\ 
       \hline
\multicolumn{6}{l}{\textit{Pretraining on ImageNet1K-1.3M}} \\
DG-Net++~\cite{zou2020joint}~(2020) & Res-50 & 64.6 & 83.1 & 22.1 & 48.4 \\
MMT~\cite{ge2020mutual}~(2020) & Res-50 & 75.6 & 83.9 & 24.0 & 50.1 \\
SpCL~\cite{ge2020self}~(2020) & Res-50 & 77.5 & 89.7 & 26.8 & 53.7 \\
C-Con~\cite{dai2021cluster}~(2021) & Res-50 & 82.4 & 92.5 & 33.4 & 60.5 \\
MCRN~\cite{wu2022multi}~(2022) & Res-50 & - & - & 32.8 & 64.4 \\ \hline
\multicolumn{6}{l}{\textit{Pretraining on LUPserson-4.2M}} \\
MoCov2~\cite{fu2021unsupervised}~(2021) & Res-50 & 85.1 & 94.4 & 28.3 & 53.8 \\
DINO~\cite{luo2021self}~(2021) & ViT-S & 88.5 & 95.0 & 43.9 & 67.7 \\
DINO-CFS~\cite{luo2021self}~(2021) & ViT-S & 89.4 & 95.4 & 47.4 & 70.8 \\
PASS~\cite{zhu2022pass}~(2022) & ViT-S & 90.2 & 95.8 & 49.1 & 72.7 \\
\rowcolor{Graylight} 
Ours & ViT-S & \textbf{90.6} & \textbf{95.9} & \textbf{53.0} & \textbf{76.1} \\
\bottomrule[1pt]
\end{tabular}
\end{threeparttable}
\end{center}
\label{tab:uda_reid}
\end{table}

\begin{table}[t!]
\small
\tabcolsep=2pt
\begin{center}
\caption{Comparison with state-of-the-art methods on USL ReID. ``MS'' and ``Mar'' denote the MSMT17 and Market1501 dataset, respectively.}
\begin{threeparttable}
\begin{tabular}{llcccc}
\toprule[1pt]
\multirow{2}{*}{\textbf{Method}} & \multirow{2}{*}{\textbf{Backbone}} & \multicolumn{2}{c}{\textbf{Mar}}  & \multicolumn{2}{c}{\textbf{MS}}  \\ \cline{3-6}
       &  & \textbf{mAP} & \textbf{Rank1} & \textbf{mAP} & \textbf{Rank1} \\ 
       \hline
\multicolumn{6}{l}{\textit{Pretraining on ImageNet1K-1.3M}} \\
MMCL~\cite{wang2020unsupervised}~(2020) & Res-50 & 45.5 & 80.3 & 11.2 & 35.4 \\
HCT~\cite{zeng2020hierarchical}~(2020) & Res-50 & 56.4 & 80.0 & - & - \\
IICS~\cite{xuan2021intra}~(2021) & Res-50 & 72.9 & 89.5 & 26.9 & 52.4 \\
C-Con~\cite{dai2021cluster}~(2021) & Res-50 & 82.6 & 93.0 & 33.1 & 63.3 \\
MCRN~\cite{wu2022multi}~(2022) & Res-50 & 80.8 & 92.5 & 31.2 & 63.6 \\ \hline
\multicolumn{6}{l}{\textit{Pretraining on LUPserson-4.2M}} \\
MoCov2~\cite{fu2021unsupervised}~(2021) & Res-50 & 84.0 & 93.4 & 31.4 & 58.8 \\
DINO~\cite{luo2021self}~(2021) & ViT-S & 87.8 & 94.4 & 38.4 & 63.8 \\
DINO-CFS~\cite{luo2021self}~(2021) & ViT-S & 88.2 & 94.2 & 40.9 & 66.4 \\
PASS~\cite{zhu2022pass}~(2022) & ViT-S & 88.5 & 94.9 & 41.0 & 67.0 \\

\rowcolor{Graylight} 
Ours & ViT-S & {88.1} & \textbf{95.3} & \textbf{48.8} & \textbf{74.2} \\
\bottomrule[1pt]
\end{tabular}
\end{threeparttable}
\end{center}
\label{tab:usl_reid}
\end{table}

\noindent \textbf{Unsupervised ReID.}
We perform comparison with previous methods on unsupervised ReID, \emph{i.e.,} UDA in Table~\ref{tab:uda_reid} and USL in Table~\ref{tab:usl_reid}.
Unsupervised ReID contains two settings, \emph{i.e.,} unsupervised domain adaptation~(UDA) and unsupervised learning~(USL).
For these two settings, our method achieves new state-of-the-art performance on most metrics.
Notably, our method outperforms the previous SOTA PASS with a large margin, \emph{i.e.,} 3.9\% mAP on Market$\rightarrow$MSMT UDA setting and 7.8\% mAP on the USL setting, respectively.
These results further verify the generalization of our designed pre-training.

\subsection{Ablation Studies}
In this section, we perform ablation studies on supervised ReID to demonstrate the effectiveness of the key components.
Unless stated, we adopt ViT-B as our backbone and conduct pre-training with 50 epochs.

\noindent \textbf{Ablation on the multi-level prediction.}
We introduce both high-level and low-level supervision during PersonMAE pre-training, and we study their impact by modifying the weighting factor $\lambda$ in Table~\ref{tab:fea_weight}.
It is observed that pixel-level regularization is necessary for convergence.
Thus we start by only adopting the pixel-level constraint, \emph{i.e.,} corresponding to the first line in Table~\ref{tab:fea_weight}~($\lambda = 0$).
Compared with this vanilla setting, adding semantic constraint further enhances the performance on downstream ReID tasks.
It can be explained that semantic constraint guides the framework to focus more on consensus on the feature space, which is in line with the ReID goal.
We further study the weighted balance between these two constraints.
It can be observed that both constraints are complementary and the best performance is achieved when $\lambda$ is set as 1.

\begin{table}[t!]
\small
\tabcolsep=13.5pt
\begin{center}
\caption{Effectiveness of multi-level prediction during pre-training. We study its effectiveness via changing the weighting factor $\lambda$. $\lambda=0$ represents that we only adopt the low-level pixel constraint.}
\begin{threeparttable}
\begin{tabular}{lcccc}
\toprule[1pt]
\multirow{2}{*}{\textbf{$\lambda$}} & \multicolumn{2}{c}{\textbf{Market1501}}  & \multicolumn{2}{c}{\textbf{MSMT17}}  \\ \cline{2-5}
       & \textbf{mAP} & \textbf{Rank1} & \textbf{mAP} & \textbf{Rank1} \\ 
       \hline 
0 & 92.6 & 96.6 & 77.1 & 90.3\\
0.5 & 93.2 & 96.8 & 77.7 & 90.5 \\
\rowcolor{Graylight} 
1.0 & \textbf{93.3} & \textbf{97.0} & \textbf{78.6} & \textbf{91.2} \\
2.0 & 93.1 & 96.9 & 78.2 & 90.8 \\
\bottomrule[1pt]
\end{tabular}
\end{threeparttable}
\end{center}
\label{tab:fea_weight}
\end{table}

\begin{table}[t!]
\small
\tabcolsep=8.6pt
\begin{center}
\caption{Effectiveness of the masking strategy during pre-training. ``Random'' and ``Block'' represent the random and block-wise masking strategies, respectively.}
\begin{threeparttable}
\begin{tabular}{lcccc}
\toprule[1pt]
\multirow{2}{*}{\textbf{Mask Strategy}} & \multicolumn{2}{c}{\textbf{Market1501}}  & \multicolumn{2}{c}{\textbf{MSMT17}}  \\ \cline{2-5}
       & \textbf{mAP} & \textbf{Rank1} & \textbf{mAP} & \textbf{Rank1} \\ 
       \hline 
Random & 92.7 & 96.9 & 76.1 & 89.8 \\
\rowcolor{Graylight} 
Block & \textbf{93.3} & \textbf{97.0} & \textbf{78.6} & \textbf{91.2} \\
\bottomrule[1pt]
\end{tabular}
\end{threeparttable}
\end{center}
\label{tab:mask_stra}
\end{table}

\noindent \textbf{Ablation on the masking strategy.}
We study two masking strategies in Table~\ref{tab:mask_stra}, \emph{i.e.,} random and block-wise masking.
The first one randomly selects the patches from the masked positions, while the latter chooses the masked patches in a block-wise manner.
Their masking ratios are both set as 75\%.
It can be observed that the block-wise masking strategy brings better performance on the downstream task than the random counterpart.
We assume that the block-wise strategy better mimics the real-world cases since common occlusion usually appears in form of a chunk.

\noindent \textbf{Ablation on the masking ratio.}
We further study the effects of the masking ratio in Table~\ref{tab:mask_ratio}.
It can be observed that the performance reaches the peak when the masking ratio is equal to 75\%.
It indicates there is much information redundancy in the pedestrian image and a suitable ratio makes the framework model its statistics well.

\begin{table}[t!]
\small
\tabcolsep=9.8pt
\begin{center}
\caption{Effectiveness of the masking ratio during pre-training.}
\begin{threeparttable}
\begin{tabular}{lcccc}
\toprule[1pt]
\multirow{2}{*}{\textbf{Mask Ratio}} & \multicolumn{2}{c}{\textbf{Market1501}}  & \multicolumn{2}{c}{\textbf{MSMT17}}  \\ \cline{2-5}
       & \textbf{mAP} & \textbf{Rank1} & \textbf{mAP} & \textbf{Rank1} \\ 
       \hline
20\% & 91.9 & 96.4 & 75.7 & 89.1  \\
40\% & 93.0 & 96.6 & 76.4 & 89.6  \\
60\% & 93.1 & 96.4 & 78.4 & 90.5  \\
70\% & 93.1 & 97.0 & 78.4 & 91.1  \\
\rowcolor{Graylight} 
75\% & \textbf{93.3} & \textbf{97.0} & \textbf{78.6} & \textbf{91.2} \\
80\% & 93.1 & 96.9 & 77.8 & 90.6\\
\bottomrule[1pt]
\end{tabular}
\end{threeparttable}
\end{center}
\label{tab:mask_ratio}
\end{table}

\begin{figure*}[t!]
\centering
\includegraphics[width=\linewidth]{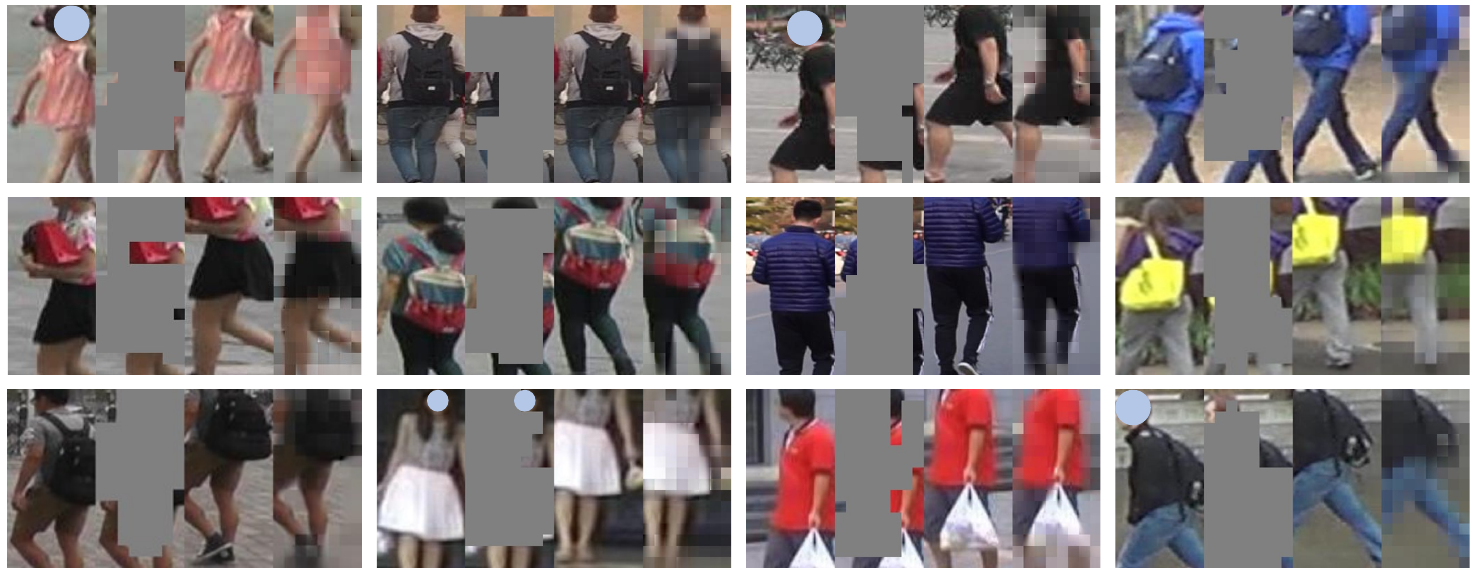}
\caption{Qualitative low-level prediction results on unseen person images. Each group contains four images, \emph{i.e.,} \textit{RegionA}, mask-distorted \textit{RegionA}, the ground-truth \textit{RegionB} and our predicted \textit{RegionB}. Our framework models the pedestrian statistics well and can reason the whole person via observing only a small partition~(25\%) of visual tokens. 
}
\label{fig:vis_recons}
\end{figure*}

\noindent \textbf{Ablation on cross-region generation.}
Cross-region generation module is utilized to simulate the cropping jittering in ReID, and we realize it with a shift operation.
We investigate the suitable shift size in Table~\ref{tab:shift_size}.
The first row corresponds to the setting when two regions are the same and the method degrades to the inner prediction case.
In this setting, although the framework is able to model the pedestrian statistics from heavy occlusion, its cross-region invariance capability is shrunk, which leads to inferior performance.
When the shift size increases, the performance gradually increases.
The best performance is achieved when the shift size is equal to 64.
Further shift size increase over 64 degrades the performance.
It can be explained that a too-large shift size increases the difficulty of the pretext task and makes pre-training not optimal.

\begin{table}[t!]
\small
\tabcolsep=10.6pt
\begin{center}
\caption{Ablation on the maximum shift size $m$ used in the cross-region generation.}
\begin{threeparttable}
\begin{tabular}{lcccc}
\toprule[1pt]
\multirow{2}{*}{\textbf{Shift Size}} & \multicolumn{2}{c}{\textbf{Market1501}}  & \multicolumn{2}{c}{\textbf{MSMT17}}  \\ \cline{2-5}
       & \textbf{mAP} & \textbf{Rank1} & \textbf{mAP} & \textbf{Rank1} \\ 
       \hline 
0 &  92.5 & 96.8 & 76.9 & 90.1 \\
32 & 92.9 & 96.9 & 77.8 & 90.9 \\
\rowcolor{Graylight} 
64 & \textbf{93.3} & \textbf{97.0} & \textbf{78.6} & \textbf{91.2} \\
96 & 92.5 & 96.6 & 76.7 & 90.1\\
\bottomrule[1pt]
\end{tabular}
\end{threeparttable}
\end{center}
\label{tab:shift_size}
\end{table}

\begin{figure}[t!]
\centering
\includegraphics[width=\linewidth]{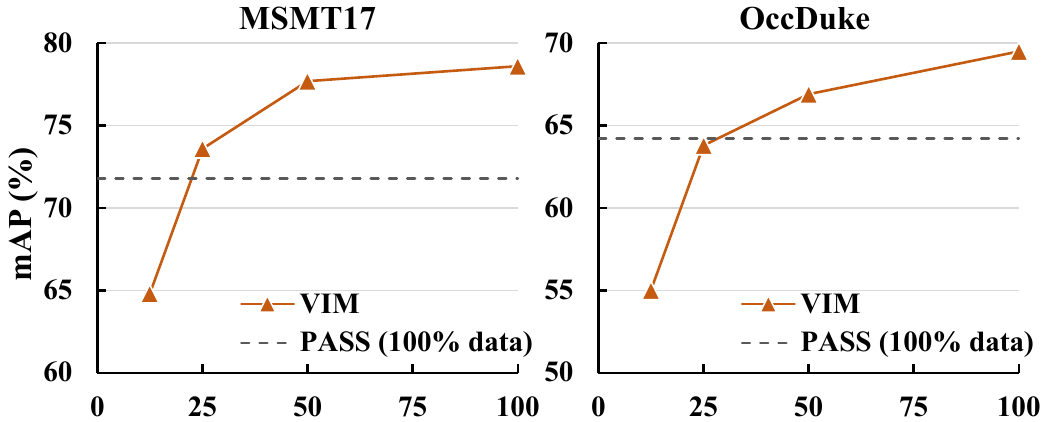}
\caption{Impact of the pre-training data scale on MSMT17 and OccDuke datasets. The x-axis and y-axis represent the pre-training data scale and mAP accuracy. Our method even outperforms PASS under the setting of a quarter of the pre-training data and half of the pre-training epochs.
}
\label{fig:vis_data_scale}
\end{figure}

\begin{figure}[t!]
\centering
\includegraphics[width=\linewidth]{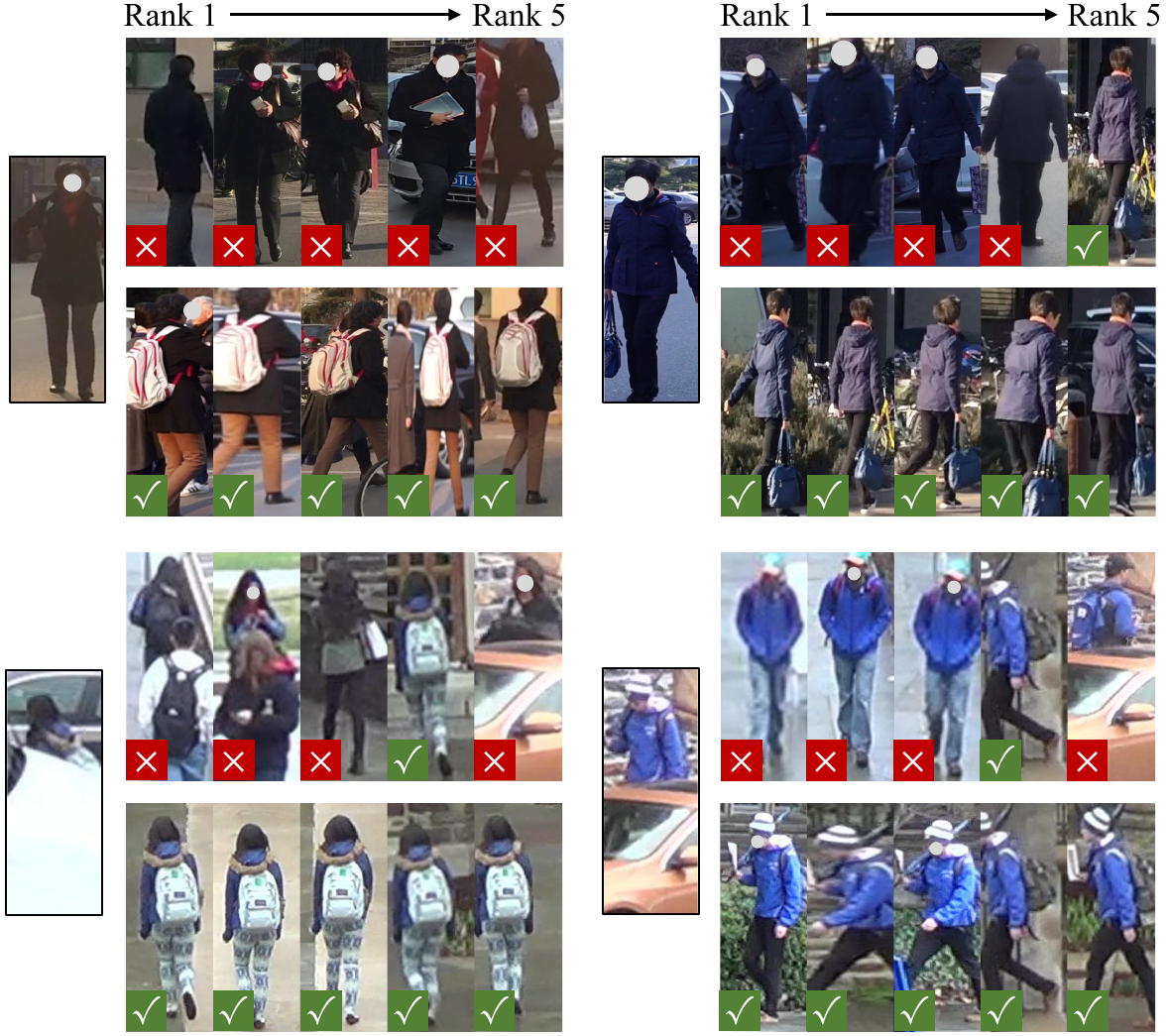}
\caption{Retrieved pedestrian images by previous SOTA PASS and our method. We list top-5 ranking results for each query. For each query, the first and second lines denote the results of PASS and ours, respectively. Green and red boxes represent the correct and false ReID, respectively.
}
\label{fig:vis_sample}
\end{figure}

\noindent \textbf{Pre-training data scale.}
We study the impact of the pre-training data scale on challenging MSMT17 and OccDuke datasets in Figure~\ref{fig:vis_data_scale}.
We randomly extract a portion of the LUPerson dataset to conduct pre-training.
Then we perform fine-tuning with the same setup.
The performance gradually improves when the involved pre-training data scale increases.
Notably, our method even outperforms previous SOTA PASS on MSMT17 by leveraging 25\% of pre-training data and 50\% of the pre-training epochs.

\noindent \textbf{Superiority over simple integration of CL and MAE.}
Another alternative to achieve the three desired properties is simply combining the objectives from CL and MAE.
To this end, we build a comparison method~(DINO+MAE), which is built based on DINO and adds MAE in its student branch.
The comparison is conducted on the ViT-S backbone.
DINO+MAE achieves 64.7(83.6)/88.0(94.9) mAP(Rank1) on MSMT17/Market1501, which is significantly worse than PersonMAE with 75.2(89.1)/92.5(96.7).

\subsection{Qualitative Analysis}
We perform qualitative visualization of our prediction results on unseen person images in Figure~\ref{fig:vis_recons}.
In each group, we show \textit{RegionA}, distorted \textit{RegionA}, ground-truth \textit{RegionB} and our predicted \textit{RegionB} in sequence.
Our framework can infer the whole cloth texture or fill the whole personal belonging well~(\emph{e.g.} bag), even with a glance at the partial region.
This strong hallucination capability may be largely attributed to the well-model pedestrian statistics during pre-training.

Furthermore, we demonstrate some retrieved pedestrian images of previous SOTA~(PASS) and our method in Figure~\ref{fig:vis_sample}.
We list top-5 ranking results for each query.
In the upper-left part, the query pedestrian wears the white schoolbag and only shows the bag straps in the query image.
However, PASS ignores this fine-grained cue and focuses on the red sweater and black coat, which leads to false retrieval.
In contrast, our framework is able to identify this detail and retrieves the pedestrian correctly.
For the bottom-right part, the query person is heavily occluded by a car, which is a highly challenging scenario.
Our framework does not rely too much on the blue coat and retrieves correctly by capturing the cues from the bag color and hat type.
These results qualitatively suggest that our method is able to satisfy the aforementioned three main properties for accurate ReID.

\section{Limitation \& Future Work}
Although our framework brings notable performance gains on the downstream ReID tasks, it still contains some failure cases.
Due to inaccurate person detection or low recording conditions, the cropped person image presents in low quality.
This factor may lead to final failure detection.
Besides, non-target pedestrian inherently introduces ambiguity on which pedestrian to match, and may cause failure retrieval.

In future work, we will explore more suitable masking strategies, which better simulate diverse occlusion characteristics or model pedestrian statistics with awareness of human parts.
We believe these directions can bring further performance improvement.
Besides, it is also desirable to explore video-based ReID tasks with the inspiration of our framework.

\section{Conclusion}
In this paper, we propose the \emph{first} pre-training framework with masked autoencoders for person ReID, namely PersonMAE.
It aims to embrace three important properties, \emph{i.e.,} multi-level awareness, occlusion robustness and cross-region invariance, to meet the challenge of ReID. 
Two regions are generated from the global pedestrian image, \emph{i.e.,} the input \textit{RegionA} and prediction target \textit{RegionB}.
\textit{RegionA} is corrupted with the masking strategy, and its unmasked parts are fed into the encoder.
Then the framework utilizes the encoder output and cross-region relationship to predict whole \textit{RegionB} on both low-level clues and high-level semantics.
Extensive experiments validate the effectiveness of our framework on a series of downstream tasks, \emph{i.e.,} supervised~(holistic and occluded ReID), unsupervised~(UDA and USL ReID) setting.
Our method achieves new state-of-the-art performance with a notable margin.

\textbf{Broader impact.}
Our research aims to meet the urgent demand for public safety.
The capability of matching a specific person across different cameras has many applications, \emph{e.g.} locating the suspect easier.
On the other hand, this task involves identifying a specific person and may have potential privacy issues if it is misused.
A variety of regulatory and technical measures have been proposed to address this issue.

{\small
\bibliographystyle{IEEEtran}
\bibliography{ref}
}

\end{document}